\documentclass{article}

\usepackage{microtype}
\usepackage{graphicx}
\usepackage{subcaption}
\usepackage{booktabs} 

\usepackage{hyperref}

\usepackage[preprint]{icml2026}

\usepackage{amsmath}
\usepackage{amssymb}
\usepackage{mathtools}
\usepackage{amsthm}

\usepackage[capitalize,noabbrev]{cleveref}

\theoremstyle{plain}

\theoremstyle{definition}

\theoremstyle{remark}

\usepackage[textsize=tiny]{todonotes}
\usepackage{listings}
\usepackage{svg}
\usepackage{xcolor}
\usepackage{multirow}
\usepackage{xurl}

\icmltitlerunning{Accurate Failure Prediction in Agents Does Not Imply Effective Failure Prevention}

\begin{document}

\twocolumn[
  \icmltitle{The Intervention Paradox: \\ Accurate Failure Prediction in Agents Does Not Imply Effective Failure Prevention}



  \icmlsetsymbol{equal}{*}
  \begin{icmlauthorlist}
    \icmlauthor{Rakshith Vasudev}{yyy}
    \icmlauthor{Melisa Russak}{yyy}
    \icmlauthor{Dan Bikel}{yyy}
    \icmlauthor{Waseem Alshikh}{yyy}
  \end{icmlauthorlist}

  \icmlaffiliation{yyy}{Writer, Inc.}

  \icmlcorrespondingauthor{Rakshith Vasudev}{rakshith@writer.com}

  \icmlkeywords{Machine Learning, ICML}

  \vskip 0.3in
]



\printAffiliationsAndNotice{}  

\begin{abstract}
Proactive interventions by LLM critic models are often assumed to improve reliability, yet their effects at deployment time are poorly understood. We show that a binary LLM critic with strong offline accuracy (AUROC 0.94) can nevertheless cause severe performance degradation, inducing a 26 percentage point (pp) collapse on one model while affecting another by near zero pp. This variability demonstrates that LLM critic accuracy alone is insufficient to determine whether intervention is safe.

We identify a disruption-recovery tradeoff: interventions may recover failing trajectories but also disrupt trajectories that would have succeeded. Based on this insight, we propose a pre-deployment test that uses a small pilot of 50 tasks to estimate whether intervention is likely to help or harm, without requiring full deployment.
Across benchmarks, the test correctly anticipates outcomes: intervention degrades performance on high-success tasks (0 to -26 pp), while yielding a modest improvement on the high-failure ALFWorld benchmark (+2.8 pp, p=0.014). The primary value of our framework is therefore identifying when not to intervene, preventing severe regressions before deployment.
\end{abstract}

\section{Introduction}

\begin{figure*}[t]
  \centering
  \includegraphics[width=0.9\linewidth]{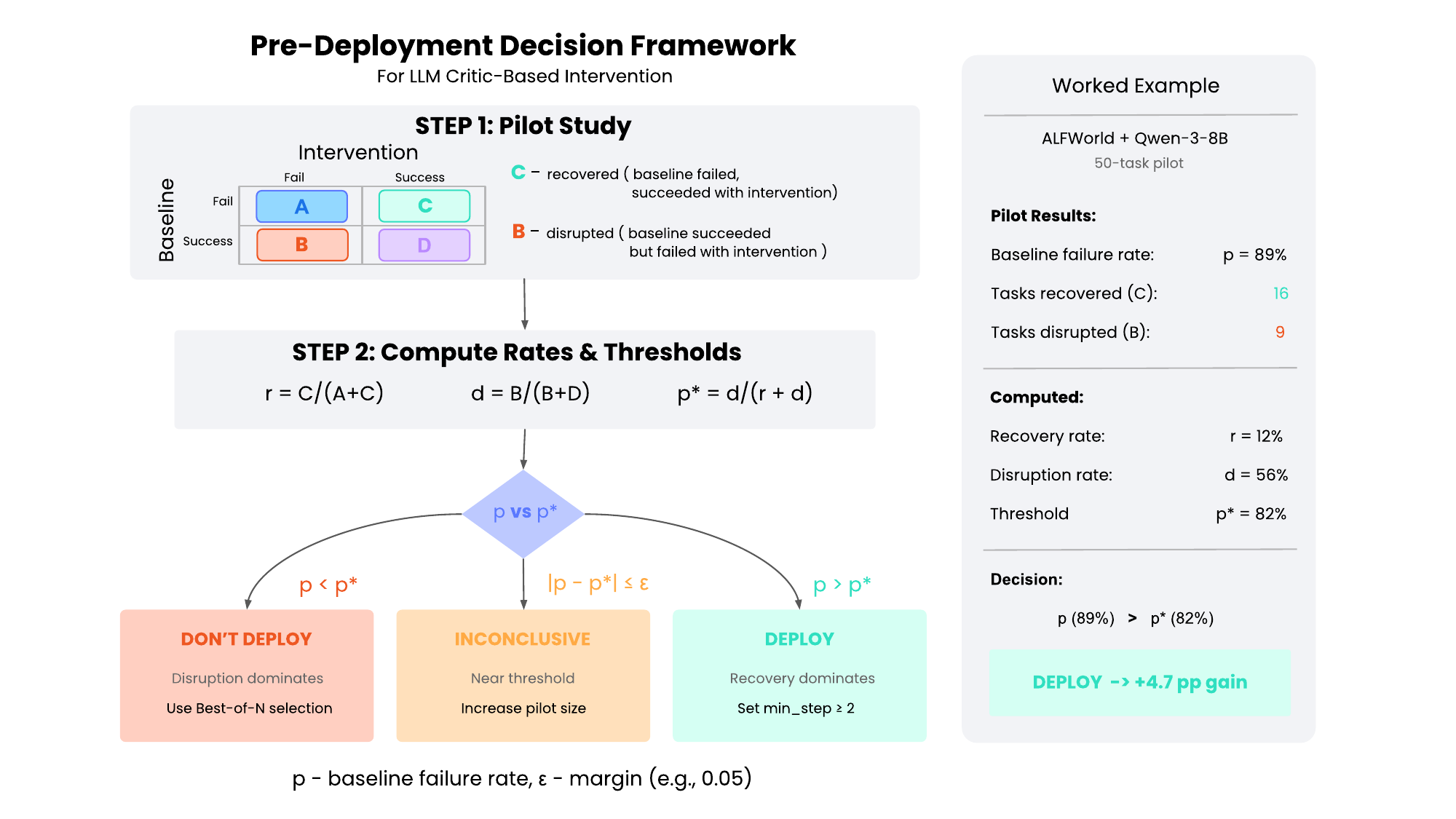}
  \caption{Decision tree illustrating the recommended deployment procedure for execution-time intervention, based on pilot estimates of failure ($p$), recovery ($r$), and disruption ($d$), and the resulting threshold $p^\star = d/(r+d)$ (left). Example calculations for ALFWorld with Qwen-3-8B are shown on the right.}
  \label{fig:decision_tree}
\end{figure*}

Large language model agents (LLM agents) are increasingly deployed for complex, multi-step tasks, where failures can be costly due to wasted computation, incorrect outputs, and degraded user experience \cite{Wang_2024, li2025review, xi2023risepotentiallargelanguage}. A common mitigation strategy is proactive intervention: a binary LLM critic model monitors execution, predicts forthcoming failure, and intervenes mid-trajectory to steer the agent back on course \cite{wu-etal-2025-automating, inan2023llamaguardllmbasedinputoutput}. Despite its intuitive appeal, intervention is not universally beneficial, and in some cases substantially degrades performance.

We formalize the effect of intervention by comparing two systems on the same tasks: \textbf{baseline} (agent only) and \textbf{intervention} (agent + critic). Running both on $N$ tasks yields a $2 \times 2$ outcome table:
\label{sec:framework}
\begin{center}
\begin{tabular}{lcc}
\toprule
& \multicolumn{2}{c}{Intervention} \\
\cmidrule(lr){2-3}
Baseline & Fail & Succeed \\
\midrule
Fail    & $A$ & $C$ \\
Succeed & $B$ & $D$ \\
\bottomrule
\end{tabular}
\end{center}
where $C$ counts \emph{recoveries} (baseline fails, intervention succeeds) and $B$ counts \emph{disruptions} (baseline succeeds, intervention fails). Let $F = A + C$ denote baseline failures and $S = B + D$ denote baseline successes. We define:
\begin{align}
p &= F / N & &\text{(baseline failure rate)}, \\
r &= C / F & &\text{(recovery rate)},\footnotemark \\
d &= B / S & &\text{(disruption rate)}.
\end{align}
\footnotetext{The rates $r$ and $d$ are undefined when $F = 0$ or $S = 0$ (all tasks succeed or fail at baseline). These edge cases correspond to trivial deployment decisions.}
The change in success rate is:
\begin{equation}
\Delta\text{Success} = p \cdot r - (1-p) \cdot d.
\end{equation}

Intervention improves performance when $\Delta\text{Success} > 0$, i.e., when $p > d/(r+d)$. Although $r$ and $d$ depend on the intervention mechanism, our experiments show their values are dominated by properties of the underlying agent, making critic accuracy a secondary factor. Consequently, the same intervention policy may improve one agent while degrading another.

Intervention trades off two forces: recovering trajectories that would have failed and disrupting trajectories that would have succeeded. Different models sit at very different points on this disruption–recovery spectrum. For some agents, a failed action can be removed or appended with little side effect; for others, the same intervention derails otherwise correct reasoning. As a result, the same LLM critic policy can mildly help one model while catastrophically harming another.

This observation has a direct implication. Even a highly accurate LLM critic does not guarantee gains. For models with a high disruption-to-recovery ratio, intervention is only beneficial when failure is already very likely---an operating regime that is rarely reachable in practice. In our experiments, this explains why an LLM critic with strong predictive signal (AUROC 0.94) consistently fails to improve performance and, for some models, causes large regressions. The magnitude of harm tracks the disruption–recovery ratio of the base agent rather than the LLM critic’s performance.

More broadly, this reframes intervention as a model-dependent systems problem rather than a prediction problem. Improving failure detection alone has a low ceiling: even perfect intervention leads to limited upside, while naïve intervention can cause substantial downside. Scaling the critic from 0.6B to 14B parameters does not improve prediction quality in our data regime (Section~3), and oracle analysis confirms that even perfect failure prediction yields at most 4 to 8 percentage point gains due to the intrinsic cost of mid-trajectory correction. The bottleneck is how the agent absorbs corrections mid-trajectory, whether it can incorporate visible failures or is destabilized by them. Any intervention scheme that ignores this tradeoff risks doing more harm than good, regardless of critic scale or accuracy.

Our contributions are threefold:
\begin{itemize}
\item We identify a disruption--recovery tradeoff governing execution-time intervention and show that the condition $p > d/(r+d)$ determines whether intervention helps or harms. While the condition itself is elementary, we demonstrate empirically that it is routinely violated in practice, leading to avoidable performance regressions of up to 26 percentage points.

\item We evaluate this framework across multiple models, benchmarks, and intervention mechanisms, showing that observed gains and regressions are well explained by differences in disruption--recovery profiles rather than LLM critic accuracy.

\item We analyze common intervention failure modes and show that when disruption dominates recovery, mid-execution control has a low ceiling and post-hoc selection provides a more effective alternative.
\end{itemize}

\section{Related Work}
A growing body of work studies how intermediate signals can improve multi-step reasoning in LLMs. Process reward models (PRMs) predict the correctness of intermediate steps and are commonly used to guide search or rerank trajectories, typically relying on step-level supervision \cite{lightman2023letsverifystepstep, uesato2022solvingmathwordproblems, setlur2024rewardingprogressscalingautomated}. 

Selection-based methods such as Best-of-$N$ decoding \cite{10.5555/3495724.3495977} and self-consistency \cite{DBLP:conf/iclr/0002WSLCNCZ23, kang2025scalablebestofnselectionlarge} evaluate completed trajectories post-hoc and choose among them without altering execution. These approaches preserve the agent’s reasoning context and have been shown to result in reliable gains, a distinction that aligns with the large gap observed between intervention and selection ceilings. \cite{marín2025capabilityceilingsautoregressivelanguage}
This approach is also used in work that employs explicit smaller models as LLM critics to rank completed trajectories by predicted outcomes, such as Prospector \cite{kim2024prospector}.

More closely related are self-correction frameworks, including Reflexion \cite{10.5555/3666122.3666499}, Self-Refine \cite{10.5555/3666122.3668141}, Reflect, Retry, Reward \cite{bensal2025reflectretryrewardselfimproving} which augment agents with feedback or retries while preserving context. Recent analysis shows that intrinsic self-correction is often unreliable and can even degrade performance, suggesting that agents struggle to accurately identify their own errors. \cite{huang2024llmscannotselfcorrect, wu-etal-2024-large}

Our contribution is orthogonal to these methods: rather than proposing a new intervention mechanism, we provide a framework for deciding \emph{whether} to deploy intervention at all, applicable to any binary LLM critic-based system. We isolate a complementary failure mode by examining when accurate external detection signals fail, or succeed, to improve outcomes once intervention is applied at execution time.

\section{Experimental Setup}

\paragraph{LLM critic model and training data.}
Motivated by deployment latency constraints, all our experiments use an LLM critic model based on Qwen3-0.6B \cite{yang2025qwen3technicalreport}, adapted via LoRA with rank 16. The LLM critic is trained on 7,636 trajectory steps collected from \texttt{smolagents} \cite{smolagents} runs on HotPotQA \cite{yang2018hotpotqa} and GAIA \cite{mialon2023gaiabenchmarkgeneralai}. Each trajectory is labeled by final task outcome (success vs.\ failure), providing a binary supervision signal aligned with end-task performance.

We partition data by unique tasks rather than by trajectories, ensuring that no task instance appears across splits. Specifically, 158 tasks are used for training (80\%), 19 for validation (10\%), and 21 for testing (10\%), with zero overlap between any pair of splits. As a result, the LLM critic is always evaluated on tasks it has never seen during training, even when training and evaluation share the same benchmark distribution. This constraint is even stricter for ALFWorld \cite{ALFWorld20}: the LLM critic is trained without any ALFWorld trajectories, making the evaluation a case of complete domain transfer.

\paragraph{LLM critic performance.}
Across all evaluated backbone LLM agents, the LLM critic achieves strong discriminative performance. Aggregated over 1,372 held-out samples, test AUROC reaches 0.936 with an F1 score of 0.963. Performance varies by agent backbone, but remains consistently high, indicating that the LLM critic learns task-agnostic failure signals rather than overfitting to a particular model family. We report AUROC and F1 results in Table \ref{tab:guard-performance}. Section \ref{section:ablations} evaluates critic scaling to 14B.

\begin{table}[h]
\small
\centering
\begin{tabular}{lccc}
\toprule
Model & AUROC & F1 & \#Samples \\
\midrule
Qwen-3-8B    & 0.982 & 0.989 & 474 \\
GLM-4.7     & 0.966 & 0.976 & 234 \\
MiniMax-M2.1& 0.897 & 0.937 & 664 \\
\midrule
Overall     & 0.936 & 0.963 & 1,372 \\
\bottomrule
\end{tabular}
\caption{LLM critic model performance on held-out tasks, aggregated by agent backbone LLM.}
\label{tab:guard-performance}
\end{table}

\paragraph{Calibration.}
For neural classifiers, including our LLM critic model, the maximum softmax probability is commonly used as a confidence score, despite being known to be miscalibrated and overconfident \cite{10.5555/3305381.3305518}. We apply a standard postprocessing Temperature Scaling to correct systematic overconfidence in raw model predictions. For each backbone LLM, a single temperature parameter $T$ is fitted by minimizing negative log-likelihood on the validation data. Unless otherwise stated, interventions are triggered when the calibrated LLM critic predicts a probability of failure above $\tau = 0.6$. Calibration substantially reduces expected calibration error (ECE) for Qwen-3-8B and GLM-4.7 (44\% and 58\% relative reductions, respectively), while having negligible effect for MiniMax-M2.1, which is already close to calibrated. Quantitative calibration results are summarized in Table \ref{tab:guard-calibration}. In all downstream experiments, we report both calibrated and uncalibrated settings to isolate the role of probability calibration from intervention mechanics.

\begin{table}[h]
\centering
\begin{footnotesize}
\begin{tabular}{lcccc}
\toprule
Model & $T$ & ECE (Before) & ECE (After) & $\Delta$ \\
\midrule
Qwen-3-8B      & 2.27 & 0.245 & 0.136 & 44\% \\
GLM-4.7       & 8.81 & 0.355 & 0.150 & 58\% \\
MiniMax-M2.1  & 1.05 & 0.133 & 0.129 & 3\%  \\
\bottomrule
\end{tabular}
\caption{Post-hoc temperature scaling calibration results. ECE is reported before and after calibration, along with the relative reduction.}
\label{tab:guard-calibration}
\end{footnotesize}
\end{table}

\paragraph{Intervention mechanisms.}
We study two deliberately simple intervention mechanisms: \texttt{ROLLBACK} and \texttt{APPEND}. In \texttt{ROLLBACK}, if the binary critic predicts failure with probability exceeding a threshold $\tau$, the agent’s most recent action is undone, the environment state is restored, and the agent is allowed to retry. In \texttt{APPEND}, the action is executed as-is, but the agent receives an additional warning message indicating that the LLM critic predicts a high likelihood of failure. 

Our choice of intervention mechanisms is intentionally minimal. First, they represent the most natural baselines a practitioner would deploy; understanding their limits is a necessary step to justifying more complex designs. Second, within our theoretical framework, any intervention is constrained by the ratio between disruption rate $d$ and recoverable benefit $r$. Simple mechanisms establish a lower bound on $d$, against which more sophisticated approaches must be measured. 

\paragraph{Experimental design.}
We combine intervention type and LLM critic calibration in a $2 \times 2$ factorial design: \texttt{ROLLBACK} vs.\ \texttt{APPEND}, crossed with calibrated vs.\ uncalibrated critic outputs. This design lets us separate the effects of calibration from the effects of the intervention itself, without adding extra complexity. In our setup, agents operate under a fixed step budget of 15 actions and an intervention budget of 3, which limits the maximum number of LLM critic interventions per episode.

\paragraph{Benchmarks.}
Experiments span three benchmarks with distinct baseline success regimes; dataset statistics and characteristics are summarized in Table \ref{tab:benchmarks}. HotPotQA \cite{yang2018hotpotqa}, a multi-hop question answering benchmark with annotated supporting facts, exhibits relatively high baseline success (51--64\%). GAIA \cite{mialon2023gaiabenchmarkgeneralai}, a benchmark for general AI assistants requiring reasoning, multimodal understanding, web interaction, and tool use, occupies a medium-success regime (19--47\%). ALFWorld \cite{ALFWorld20} represents a high-failure setting (5.8--14.7\%) and consists of a deterministic household robotics simulation. This diversity of regimes allows us to stress-test interventions both in settings where failures are rare and where they are the norm.

\begin{table}[h]
\begin{footnotesize}
\centering
\begin{tabular}{lccc}
\toprule
Benchmark & $\#$ Tasks & Baseline Success & Regime \\
\midrule
HotPotQA  & 100 & 51--70\%   & High-success \\
GAIA      & 30  & 19--47\%   & Medium-success \\
ALFWorld  & 202 & 5.8--14.7\% & Low-success \\
\bottomrule
\end{tabular}
\caption{Benchmarks used in our evaluation, including the number of tasks, baseline success rates, and difficulty regime.}
\label{tab:benchmarks}
\end{footnotesize}
\end{table}

\section{Main Results}

The disruption--recovery framework makes a testable prediction: intervention should become beneficial once the baseline failure rate exceeds $d/(r+d)$. We test this prediction in three distinct baseline success regimes: HotPotQA (high-success), GAIA (medium-success), and ALFWorld (low-success). Across all settings, we report mean success over 2 or 3 random seeds, with task-level bootstrap confidence intervals; results are summarized in Table~\ref{tab:interventions_vertical}.

\paragraph{High-success regime: HotPotQA.}
On HotPotQA, where baseline success ranges from 57--70\% depending on the backbone LLM, intervention consistently fails to improve performance and often causes regressions. For Qwen-3-8B, interventions do not produce measurable gains over the baseline, with the best setting still underperforming by roughly 2-3 pp. For GLM-4.7, effects are neutral-to-mildly negative (0 to $\approx$4 pp). In contrast, MiniMax-M2.1 exhibits extreme sensitivity: all intervention variants reduce success by 25--30 pp, far outside baseline uncertainty. These effects are statistically significant and persist across calibration and mechanism choices.

\begin{table*}[t]
\small
\centering
\begin{tabular}{lcccccc}
\toprule
Model & Baseline [$\%$]& Uncal+ Roll [$\%$] & Cal+Roll [$\%$] & Uncal+App [$\%$] & Cal+App [$\%$] & Best $\Delta$ [pp]\\
\midrule
\multicolumn{5}{l}{\textbf{HotPotQA}} \\
\midrule
Qwen-3-8B
& 57.0 (55.0, 58.0)
&53.3 (50.0, 56.0)& 54.7 (54.0, 56.0)
&51.3 (48.0, 53.0)& 49.7 (48.0, 52.0)
& \textcolor{red}{-2.3} \\

GLM-4.7
& 70.3 (66.0, 74.0)
&70.3 (69.0, 72.0)& 68.7 (68.0, 69.0)
&67.3 (64.0, 70.0)& 66.7 (63.0, 71.0)
& 0.0 \\

MiniMax-M2.1
& 64.0 (61.0, 69.0)
&38.0 (35.0, 40.0)& 38.5 (35.0, 42.0)
&34.7 (34.0, 35.0)& 36.7 (34.0, 38.0)
& \textcolor{red}{-25.5} \\
\midrule
\multicolumn{5}{l}{\textbf{GAIA}} \\
\midrule
Qwen-3-8B
& 18.9 (16.7, 20.0)
& 14.5 (10.0, 16.7)& 12.2 (10.0, 16.7)
& 8.9 (3.3, 13.3)& 10.0 (6.7, 13.3)
& \textcolor{red}{-4.4} \\

GLM-4.7
& 34.4 (23.3, 43.3)
&23.3 (20.0, 26.7)& 27.8 (26.7, 30.0)
&23.3 (16.7, 30.0)& 31.1 (23.3, 36.7)
& \textcolor{red}{-4.4} \\

MiniMax-M2.1
& 46.7 (43.3, 50.0)
&13.3 (6.7, 16.7)& 16.7 (13.3, 20.0)
&12.2 (3.3, 26.7)& 14.4 (13.3, 16.7)
& \textcolor{red}{-30.0} \\

\midrule
\multicolumn{5}{l}{\textbf{ALFWorld}} \\
\midrule

Qwen-3-8B
&5.8 (3.8, 8.1)&7.9 (5.4, 11.9)&6.9 (4.5, 9.9)&8.6 (5.6, 11.9)&7.8 (5.3, 10.6)&\textcolor{teal}{+2.8} \\

GLM-4.7
&14.7 (13.9, 16.3)&15.8 (14.9, 16.3)&15.0 (14.4, 15.8)&15.3 (13.4, 16.8)&13.4 (11.4, 14.4)&\textcolor{teal}{+1.1} \\

MiniMax-M2.1
&16.1 (15.3, 16.8)&16.6 (15.3, 17.8)&15.8 (15.3, 16.3)&15.3 (14.4, 15.8)&16.5 (15.8, 16.8)&\textcolor{teal}{+0.5} \\

\bottomrule
\end{tabular}
\caption{Performance on HotPotQA, GAIA, and ALFWorld under baseline and intervention mechanisms \texttt{ROLLBACK} (Roll) and \texttt{APPEND} (App), reporting both uncalibrated (Uncal) and calibrated (Cal) results. 95\% confidence intervals are shown in parentheses. Best $\Delta$ denotes the largest change relative to baseline.}
\label{tab:interventions_vertical}
\end{table*}

\paragraph{Medium-success regime: GAIA.} GAIA, despite being more challenging overall, exhibits the same qualitative pattern. No intervention condition outperforms the baseline for any model. Qwen-3-8B and GLM-4.7 show moderate degradations (roughly $-4$ to $-13$ pp), while MiniMax-M2.1 again suffers catastrophic losses exceeding 30\%. Although GAIA confidence intervals are wider due to the smaller task set, the direction of the effect is unambiguous and mirrors HotPotQA.

Taken together, these results demonstrate that strong LLM critic discrimination alone is insufficient. Even with high AUROC, intervention degrades performance whenever the baseline failure rate is below the disruption--recovery threshold. Moreover, model sensitivity varies by more than an order of magnitude: the same intervention that mildly harms Qwen-3-8B or GLM-4.7 catastrophically destabilizes MiniMax-M2.1, consistent with large differences in disruption rate $d$.

\paragraph{Low-success regime: ALFWorld.}
For ALFWorld, a deterministic household robotics benchmark that differs substantially from QA tasks, the LLM critic is evaluated under zero-shot transfer: the LLM critic weights are frozen and only the temperature scaling is fitted based on the ALFWorld validation data.

A 50-task pilot study establishes a baseline success of $10.7\%\pm1.9\%$ for Qwen-3-8B, corresponding to a failure rate of approximately 89\%. Estimating
intervention outcomes results in a recovery rate $r\approx12\%$ and a disruption rate $d\approx56\%$, implying a threshold $p^\star\approx82\%$. Since the observed failure rate exceeds this threshold ($89\% > 82\%$), our framework predicts a positive net effect. Indeed, the results confirm our claim: all standard intervention mechanisms in the pilot satisfy the necessary condition of more recovered than disrupted episodes, and all lead to positive changes. 

On the pilot, \texttt{ROLLBACK} achieves the largest improvement (+4.7 pp), while the full 202-task evaluation confirms uncalibrated \texttt{APPEND} as the best statistically significant gain (+2.8 pp, $p=0.014$), while \texttt{ROLLBACK} results in a larger absolute improvement of +4.7 pp. Importantly, no intervention causes performance degradation in this regime. Although the absolute gains are modest, their magnitude is consistent with the theoretical bounds implied by the estimated $r$, $d$, and $p$.

\paragraph{Mechanism-level interpretation.}
In every ALFWorld case with non-negative gains, the number of recovered episodes exceeds the number disrupted ($d/r < 1$). Conversely, in HotPotQA and GAIA, all harmful settings violate this inequality. This simple accounting fully explains the sign of intervention effects across benchmarks, models, and mechanisms.

Calibration interacts with this tradeoff in a regime-dependent manner. On ALFWorld, temperatures fitted on QA data suppress intervention too aggressively, reducing recovery opportunities early in trajectories. As a result, uncalibrated LLM critics outperform calibrated ones in the low-success regime, the opposite of what is observed in high-success settings. This highlights that calibration is not universally beneficial: it must be matched to the underlying recovery dynamics.

\paragraph{Model sensitivity.}
Finally, we summarize model-dependent sensitivity to intervention. The observed effects do not align with model scale: across a wide range of parameter counts, larger models are not consistently more robust to intervention than smaller ones. MiniMax-M2.1 exhibits extreme disruption, leading to catastrophic losses in both HotPotQA and GAIA, despite its large scale. Two factors explain this: calibration reduces MiniMax's intervention rate by only 3\% (vs.\ 71\% for GLM), and MiniMax recovers from only 12\% of interventions (vs.\ 25\% for GLM), yielding a disruption-to-recovery ratio of 7.3:1 (Appendix~\ref{app:sensitivity}). In contrast, Qwen-3-8B and GLM-4.7 are substantially more robust, experiencing only moderate harm in unfavorable regimes and modest gains in favorable ones.

In summary, execution-time intervention exhibits strong dependence on both task regime and model. In high-success settings, it consistently degrades performance, whereas in low-success settings it can lead to small but statistically meaningful gains. Across all experiments, the disruption–recovery framework correctly predicts the direction of these effects, including cross-domain transfer to ALFWorld.

\fbox{
  \parbox{\dimexpr\linewidth-2\fboxsep-2\fboxrule}{
  
\paragraph{Takeaway.}
The dominant limitation is not failure prediction, but the agent’s ability to absorb mid-trajectory corrections. Interventions that ignore this disruption–recovery tradeoff $(d/r)$ risk inducing large and avoidable regressions.
}}

\section{Ablations}
\label{section:ablations}
This section examines whether the negative results in high-success regimes arise from (i) sub-1B LLM critic choice (ii) thresholding, (iii) the wording of the feedback text itself, or (iv) the use of a learned LLM critic rather than simple heuristics. 

\subsection{Scaling the LLM critic}
A natural question is whether a larger critic would reduce intervention failures. We trained a 14B-parameter critic (Qwen3-14B, LoRA) under four configurations spanning learning rates, ranks, and regularization strengths (Table~\ref{tab:guard-scale}). The 14B~v1 variant uses identical hyperparameters to the 0.6B critic (rank~16, lr=$2{\times}10^{-4}$, 3~epochs), isolating model scale as the only variable. Despite a 23$\times$ increase in parameters, 14B~v1 achieves lower overall AUROC (0.905 vs.\ 0.936). The best 14B configuration (v2: rank~16, lr=$5{\times}10^{-5}$, label smoothing~0.1) reaches 0.927, still below the 0.6B baseline. Higher-capacity configurations overfit: rank~64 (v3) collapses to 0.752. With approximately 4{,}000 training examples drawn from two benchmarks, training data diversity---not model capacity---is the binding constraint. These results indicate that scaling the critic does not improve failure prediction in our data regime, consistent with the oracle analysis showing that even perfect prediction yields limited intervention gains (Section~5.5).

\begin{table}[h]
\centering
\begin{footnotesize}
\begin{tabular}{lccccc}
\toprule
Model & 0.6B & 14B v1 & 14B v2 & 14B v3 & 14B v4 \\
\midrule
Config & r16 & r16 & r16 & r64 & r8 \\
       & lr=2e-4 & lr=2e-4 & lr=5e-5 & lr=5e-5 & lr=2e-5 \\
\midrule
Qwen    & 0.982 & 0.986 & 0.980 & 0.918 & 0.957 \\
GLM     & 0.966 & 0.949 & 0.948 & 0.509 & 0.929 \\
MiniMax & 0.897 & 0.879 & 0.891 & 0.700 & 0.857 \\
\midrule
Overall & \textbf{0.936} & 0.905 & 0.927 & 0.752 & 0.901 \\
\bottomrule
\end{tabular}
\end{footnotesize}
\caption{AUROC by critic model size and LoRA configuration. The 0.6B critic outperforms all 14B variants. v1 uses identical hyperparameters to the 0.6B critic, isolating model scale. v2--v4 use label smoothing~0.1; v4 adds weight decay~0.05, dropout~0.15, and trains for 2~epochs instead of~3.}
\label{tab:guard-scale}
\end{table}

\subsection{Threshold sensitivity}
We first test whether the observed regressions are driven by a potentially suboptimal intervention threshold $\tau = 0.6$. On Qwen-3-8B / HotPotQA, sweeping $\tau$ shows a non-monotonic response: low thresholds over-intervene, while very high thresholds under-intervene but still trigger on occasional false positives. The best-performing threshold is $\tau=0.7$, but it still underperforms the no-intervention baseline by approximately 3 pp. Full sweep results are reported in Table~\ref{tab:threshold-sweep}.

\begin{table}[h]
\small
\centering
\begin{tabular}{lcc}
\toprule
$\tau$ & Success [\%] & $\Delta$ vs.\ baseline\\
\midrule
0.4 & 51 & \textcolor{red}{-6} \\
0.5 & 43 & \textcolor{red}{-14} \\
0.6 & 51 & \textcolor{red}{-6} \\
0.7 & 54 & \textcolor{red}{-3} \\
0.8 & 52 & \textcolor{red}{-5} \\
0.9 & 48 & \textcolor{red}{-9} \\
\bottomrule
\end{tabular}
\caption{Threshold sweep for calibrated \texttt{ROLLBACK} on Qwen-3-8B / HotPotQA. The best-performing threshold remains below the no-intervention baseline (57\%), suggesting that threshold choice alone does not explain the observed regressions.}

\label{tab:threshold-sweep}
\end{table}

\fbox{
  \parbox{\dimexpr\linewidth-2\fboxsep-2\fboxrule}{
  
\paragraph{Takeaway.}
Choosing $\tau=0.6$ is not the source of the negative result; even the optimal $\tau$ cannot recover the baseline performance. This is consistent with the disruption--recovery condition: adjusting $\tau$ rebalances intervention frequency but does not change the underlying $d/r$ of the mechanism.

}}

\subsection{Role of feedback content}
We examine whether degradation is due to the feedback message itself (e.g., confusing or distracting the agent) rather than to intervention. First, we compare ``visible feedback'' (append an explicit warning) against a silent variant that performs the same control action without adding text.

The effect is model-dependent. For Qwen-3-8B, removing feedback improves \texttt{APPEND}-style intervention (the warning appears to distract), while for GLM it has a negative effect (the warning provides useful signal). This indicates that intervention outcomes are shaped primarily by the agent’s response behavior.

\begin{table}[h]
\small
\centering
\begin{tabular}{l l c c}
\toprule
Model & Condition & With Feedback & No Feedback \\
\midrule
\multirow{2}{*}{Qwen-3-8B}
 & Cal+Roll & 54.7\% & 53.7\% \\
 & Cal+App   & 49.7\% & 52.0\% \\
\midrule
\multirow{2}{*}{GLM-4.7}
 & Cal+Roll & 70.0\% & 68.0\% \\
 & Cal+App   & 71.0\% & 68.7\% \\
\bottomrule
\end{tabular}
\caption{Effect of feedback on HotPotQA performance for Qwen-3-8B and GLM-4.7 under calibrated intervention settings.}
\label{tab:feedback-hotpotqa}
\end{table}

  


Secondly, we test whether richer feedback can reduce intervention harm by increasing recoveries (raising $r$) without increasing disruptions (raising $d$). When the LLM critic triggers, we replace the generic warning with a short, context-aware explanation generated by the agent model itself, conditioned on the task, trajectory, and flagged action.

We report the key results in Table~\ref{tab:llm_feedback}. For Qwen-3-8B the LLM feedback slightly improves the \texttt{APPEND} mechanism but does not lead to net improvement over baseline. For MiniMax-M2.1, contextual feedback substantially \emph{worsens} outcomes.

\begin{table}[h]
\scriptsize
\centering
\begin{tabular}{lccccc}
\toprule
Model & Baseline [\%] & Feedback [\%] & LLM Feedback [\%]\\
\midrule
Qwen-3-8B & 57.0 & 49.7 & 54.0 \\
MiniMax-M2.1 & 64.0 & 36.7 & 24.7 \\
\bottomrule
\end{tabular}
\caption{Effect of contextual (LLM-generated) feedback when the LLM critic triggers. Values are success rates (\%). Richer feedback can reduce harm for some models (Qwen-3-8B) but can also amplify disruption (MiniMax-M2.1).}
\label{tab:llm_feedback}
\end{table}

\fbox{
  \parbox{\dimexpr\linewidth-2\fboxsep-2\fboxrule}{
\paragraph{Takeaway.}
Feedback sophistication is not a monotone improvement knob: it can raise disruption as easily as it raises recovery.}}




\subsection{Learned vs.\ heuristic intervention policies}
A natural question is whether the learned LLM critic provides value beyond simple heuristics. We compare against (i) random intervention at fixed rates, (ii) a fixed-rate policy matched to the LLM critic’s intervention frequency, and (iii) a late-only rule that only allows intervention after step 5. Results are shown in Table~\ref{tab:heuristic-comparison}. All intervention variants remain below the baseline on Qwen / HotPotQA, including the learned LLM critic at its best threshold $\tau$.

\begin{table}[h]
\centering
\small
\begin{tabular}{l c}
\toprule
Policy & Cal+Roll [\%] \\
\midrule
  Baseline (no LLM critic) 
  & 57.0 \\
  \midrule
  A: Intervene at random 20\% steps
  & 48.3 \\
  B: Intervene at random 50\% of steps
  & 41.7 \\
  C: Intervene at LLM critic match rate
  & 49.0 \\
  D: Intervene only after step 5 
  & 53.0 \\
  E: Learned LLM critic $\tau=0.6$
  & 51.0 \\
  F: Learned LLM critic $\tau=0.7$
  & 54.0 \\
\bottomrule
\end{tabular}
\caption{Comparison of learned LLM critic interventions to simple heuristic policies on Qwen-3-8B / HotPotQA.}
\label{tab:heuristic-comparison}
\end{table}

Two comparisons are particularly informative. First, the learned LLM critic (E) is broadly comparable to a matched-rate heuristic (C), suggesting that the bulk of harm comes from intervention itself rather than from poor triggering. Second, the policy with late intervention (D) performs similarly to the best LLM critic variants (F), indicating that a coarse ``avoid early intervention'' heuristic captures most of the LLM critic’s practical benefit in this regime.

\fbox{
  \parbox{\dimexpr\linewidth-2\fboxsep-2\fboxrule}{
\paragraph{Takeaway.}
In high-success settings, the learned LLM critic does not deliver meaningful gains over simple rules; the limiting factor is the intervention mechanism’s disruption cost.
}}

\subsection{Oracle ceiling: limits of intervention.}
\label{ceil}
We estimate an oracle upper bound by intervening only on episodes that fail without intervention. Even under this idealized setting, the improvement from mid-execution intervention is modest (3--8 pp across models). 

For comparison, an oracle post-hoc selection strategy (e.g., Best-of-2 with perfect ranking) gives substantially larger gains (11--17 pp), since it operates on completed trajectories and avoids mid-execution disruption. 

HotPotQA results for both oracle intervention and oracle post-hoc Best-of-2 selection are summarized in Table~\ref{tab:oracle_ceiling_comparison}.

\begin{table}[h]
\centering
\scriptsize
\begin{tabular}{lccccc}
\toprule
Model & Baseline [\%] & LLM critic Ceil [\%] ($\Delta$) & Bo2 Ceil [\%] ($\Delta$) \\
\midrule
Qwen-3-8B    & 57.0 & 64.7 (+7.7)& 68.0 (+11.0) \\
GLM-4.7     & 70.3 & 75.0 (+4.7) & 77.0 (+6.7) \\
MiniMax-M2.1  & 64.0 & 68.0 (+4.0) & 75.0 (+11.0) \\
\bottomrule
\end{tabular}
\caption{HotPotQA ceiling comparison between oracle LLM critic-based intervention for Cal: \texttt{ROLLBACK} (with perfect failure prediction) and oracle best-of-two selection (Bo2; perfect trajectory ranking). Improvements are in percentage points (pp) over the baseline.}
\label{tab:oracle_ceiling_comparison}
\end{table}

  
\subsection{Summary}
Across ablations, we identify three consistent patterns. (i) Threshold tuning changes intervention frequency but does not fix negative net effects when $p$ is below the disruption--recovery threshold. (ii) Feedback and timing interact strongly with the agent, producing model-specific responses that are not predictable from LLM critic accuracy. (iii) Simple heuristics (e.g., avoiding early intervention) match the best learned-LLM critic behavior in high-success regimes, implying that the core bottleneck is the intervention mechanism’s disruption cost rather than sophisticated failure prediction.

\section{Early-Step Intervention as a Dominant Failure Mode}
Analysis of matched baseline and intervention runs shows that nearly all harm events arise from interventions applied at steps 0--1, where the baseline agent would have succeeded immediately. On HotPotQA, all observed regressions for both \texttt{ROLLBACK} and \texttt{APPEND} fall into this category, indicating that performance degradation is driven by disrupting already-correct trajectories rather than by errors in long-horizon reasoning.

Early interventions act as strong negative signals that can destabilize sensitive models. In practice, a single rollback at step 0 often causes the agent to abandon a correct answer and switch strategies (e.g., unnecessary search), triggering repeated interventions and exhausting the intervention budget. Once this cascade occurs, agents frequently fail to recover even with substantial remaining step budget. This mechanism explains the large cross-model differences observed earlier and motivates enforcing a minimum-step constraint (e.g., \texttt{min\_step}$\ge2$) to prevent early-step intervention.

\section{Practical Guidelines}
Our results suggest a simple deployment rule: \emph{estimate the disruption--recovery profile of the target agent in the target domain before enabling proactive intervention}. The key quantity is the threshold $p^\star = d/(r+d)$, below which intervention is expected to be net harmful. Figure~\ref{fig:decision_tree} illustrates this guideline as a simple decision tree.

\paragraph{Pilot before deployment.}
Before full deployment, run a small pilot study (e.g., 50--100 tasks) using the chosen intervention mechanism (we use calibrated \texttt{ROLLBACK}). This pilot is used to estimate the baseline failure rate $p$, the recovery rate $r$, and the disruption rate $d$ by comparing task outcomes with the intervention enabled versus disabled under matched conditions.
Deploy only if $p$ exceeds $p^\star= d/(r+d)$ by a safety margin.

\paragraph{Avoid early-step interventions.}
Empirically, most regressions in high-success settings arise from interventions applied at steps 0--1, where trajectories are often already correct. Imposing a minimum-step constraint (e.g., \texttt{min\_step}$\ge 2$) reduces these failures and recovers a non-trivial fraction of avoidable harm. 

\paragraph{When $d/r>1$, prefer selection over intervention.}
If $d/ r>1$, then $p^\star>0.5$ and intervention can only help when the agent is more likely to fail than succeed. In this regime, mid-execution control is typically a poor trade: even oracle intervention has a low ceiling (single-digit points), while oracle Best-of-$2$ selection exhibits substantially larger headroom without disruption. Thus, when additional compute is available, running multiple trajectories and selecting post-hoc is often the safer and more effective alternative.

\section{Limitations}

\paragraph{Critic model scale.}
All intervention experiments use a 0.6B-parameter critic. We also trained a 14B critic under four LoRA settings (Table~\ref{tab:guard-scale}), but the 0.6B model achieved higher AUROC (0.936 vs.\ 0.927 best at 14B). This suggests that data diversity, rather than model capacity, is the main limitation in our setting. Oracle analysis (Section~\ref{ceil}) shows that even perfect failure prediction would yield only a 4--8 pp gain, further indicating that critic scale is not the dominant bottleneck. Future work should test whether much larger or frontier-scale critics can directly reduce disruption rates.

\paragraph{Intervention mechanism complexity.}
We study two intentionally simple mechanisms (\texttt{ROLLBACK} and \texttt{APPEND}). More targeted methods—such as step-specific corrections or planning-aware backtracking—may reduce disruption. Our mechanisms set a lower bound on $d$ for comparison with more complex designs. The framework makes this explicit: any new mechanism must achieve $d/(r+d)$ below the target failure rate to be viable.

\paragraph{Benchmark and agent coverage.}
Results cover three benchmarks and three agent backbones within a single framework (\texttt{smolagents}). Disruption–recovery behavior may differ in other frameworks or task domains. We recommend using the pilot procedure as a domain-specific diagnostic, rather than treating our $r$ and $d$ estimates as universal.

\paragraph{Statistical power on positive results.}
The ALFWorld gain (+2.8 pp, $p=0.014$) is close to the detection limit (±4 pp at 80\% power for 202 tasks × 3 seeds). We treat this as directional evidence rather than a precise estimate. By contrast, the negative results in high-success regimes are well above the detection limit and are statistically clear.

\paragraph{Pilot transfer.}
The pilot-based deployment test is validated within-distribution (ALFWorld pilot predicts ALFWorld outcomes). Cross-distribution transfer of pilot estimates, such as estimating $d$ on QA tasks and deploying on coding tasks, is untested and likely unreliable given the strong model- and domain-dependence we observe.

\section{Conclusion}

This work examines execution-time intervention for LLM agents and identifies a fundamental tradeoff between disruption and recovery that governs when intervention is beneficial. Across multiple models, benchmarks, and intervention mechanisms, outcomes are driven less by the accuracy of failure prediction than by how the underlying agent responds to being corrected mid-trajectory.

Intervention consistently degrades performance in high-success regimes, with effects ranging from mild regressions to catastrophic collapse depending on the agent.

In contrast, in low-success regimes, intervention can lead to modest but reliable gains. A pilot-based estimation of recovery and disruption rates correctly predicts this regime shift, and full-scale evaluation confirms positive effects without significant harm. However, even in favorable regimes, gains remain bounded (on the order of 1--3 pp), as shown by an oracle LLM critic analysis, indicating a low ceiling for critic-based intervention.

We propose a disruption–recovery framework that captures these effects in a simple condition: expected improvement requires $p > d/(r+d)$, where $r$ and $d$ denote recovery and disruption rates. This criterion explains variation across domains and agents.

These results translate into concrete deployment guidelines: intervention should not be deployed by default but evaluated via a small pilot estimating $r$ and $d$; early-step intervention should be avoided; and when disruption outweighs recovery ($d/r > 1$), post-hoc selection dominates mid-execution control.

Overall, LLM critic-based intervention is better viewed as a model-dependent control problem than a prediction task. Gains from improved critic accuracy are limited unless agents can incorporate corrections without destabilizing behavior, implying limited upside and meaningful risk outside high-failure regimes.

\section*{Impact Statement}
This work provides deployment guidelines for execution-time intervention in LLM agents,
  helping practitioners identify when such intervention will help versus harm.
  The primary benefit is \emph{avoiding unnecessary failures}. LLM critic models with high
  offline accuracy are often assumed safe to deploy, yet we show they can degrade
  performance by up to 26 percentage points in unfavorable regimes. Our pilot-based
  framework enables teams to detect these failure modes before production deployment.
  A secondary benefit is \emph{resource efficiency}. By characterizing when intervention
  cannot help, practitioners can redirect compute toward alternatives (e.g., post-hoc
  selection) that our analysis shows have higher ceilings in those settings.
  This work does not introduce new agent capabilities. It provides diagnostic tools
  for safer deployment of existing systems, complementing rather than replacing
  existing approaches such as self-correction and process supervision.

\bibliography{references}
\bibliographystyle{icml2026}

\newpage
\appendix
\onecolumn

\section{Per-Seed Results}
\label{app:per-seed}

Tables~\ref{tab:qwen-hotpot-seeds}--\ref{tab:glm-alf-seeds} report per-seed success rates for all model--benchmark combinations. Mean values correspond to those in Table~\ref{tab:interventions_vertical}.

\subsection{HotPotQA}

\begin{table}[h]
\centering
\begin{tabular}{lcccc}
\toprule
Condition & Seed 42 & Seed 123 & Seed 456 & Mean \\
\midrule
Baseline       & 58\% & 58\% & 55\% & 57.0\% \\
Uncal+Roll     & 54\% & 56\% & 50\% & 53.3\% \\
Cal+Roll       & 56\% & 54\% & 54\% & 54.7\% \\
Uncal+App      & 53\% & 53\% & 48\% & 51.3\% \\
Cal+App        & 52\% & 48\% & 49\% & 49.7\% \\
\bottomrule
\end{tabular}
\caption{Per-seed accuracy for Qwen-3-8B on HotPotQA.}
\label{tab:qwen-hotpot-seeds}
\end{table}

\begin{table}[h]
\centering
\begin{tabular}{lcccc}
\toprule
Condition & Seed 42 & Seed 123 & Seed 456 & Mean \\
\midrule
Baseline       & 74\% & 71\% & 66\% & 70.3\% \\
Uncal+Roll     & 72\% & 69\% & 70\% & 70.3\% \\
Cal+Roll       & 69\% & 68\% & 69\% & 68.7\% \\
Uncal+App      & 64\% & 70\% & 68\% & 67.3\% \\
Cal+App        & 71\% & 66\% & 63\% & 66.7\% \\
\bottomrule
\end{tabular}
\caption{Per-seed accuracy for GLM-4.7 on HotPotQA.}
\label{tab:glm-hotpot-seeds}
\end{table}

\begin{table}[h]
\centering
\begin{tabular}{lcccc}
\toprule
Condition & Seed 42 & Seed 123 & Seed 456 & Mean \\
\midrule
Baseline       & 61\% & 69\% & 62\% & 64.0\% \\
Uncal+Roll     & 40\% & 35\% & 39\% & 38.0\% \\
Cal+Roll       & 35\% & 42\% & 37\% & 38.0\% \\
Uncal+App      & 35\% & 34\% & 35\% & 34.7\% \\
Cal+App        & 34\% & 38\% & 38\% & 36.7\% \\
\bottomrule
\end{tabular}
\caption{Per-seed accuracy for MiniMax-M2.1 on HotPotQA.}
\label{tab:minimax-hotpot-seeds}
\end{table}

\subsection{GAIA}

\begin{table}[h]
\centering
\begin{tabular}{lcccc}
\toprule
Condition & Seed 42 & Seed 123 & Seed 456 & Mean \\
\midrule
Baseline     & 16.7\% & 20.0\% & 20.0\% & 18.9\% \\
Uncal+Roll   & 16.7\% & 10.0\% & 16.7\% & 14.5\% \\
Cal+Roll     & 16.7\% & 10.0\% & 10.0\% & 12.2\% \\
Uncal+App    & 3.3\%  & 10.0\% & 13.3\% & 8.9\%  \\
Cal+App      & 10.0\% & 6.7\%  & 13.3\% & 10.0\% \\
\bottomrule
\end{tabular}
\caption{Per-seed accuracy for Qwen-3-8B on GAIA.}
\label{tab:qwen-gaia-seeds}
\end{table}

\begin{table}[h]
\centering
\begin{tabular}{lcccc}
\toprule
Condition & Seed 42 & Seed 123 & Seed 456 & Mean \\
\midrule
Baseline     & 36.7\% & 43.3\% & 23.3\% & 34.4\% \\
Uncal+Roll   & 20.0\% & 23.3\% & 26.7\% & 23.3\% \\
Cal+Roll     & 26.7\% & 30.0\% & 26.7\% & 27.8\% \\
Uncal+App    & 23.3\% & 30.0\% & 16.7\% & 23.3\% \\
Cal+App      & 36.7\% & 23.3\% & 33.3\% & 31.1\% \\
\bottomrule
\end{tabular}
\caption{Per-seed accuracy for GLM-4.7 on GAIA.}
\label{tab:glm-gaia-seeds}
\end{table}

\begin{table}[h]
\centering
\begin{tabular}{lcccc}
\toprule
Condition & Seed 42 & Seed 123 & Seed 456 & Mean \\
\midrule
Baseline     & 50.0\% & 46.7\% & 43.3\% & 46.7\% \\
Uncal+Roll   & 6.7\%  & 16.7\% & 16.7\% & 13.3\% \\
Cal+Roll     & 13.3\% & 16.7\% & 20.0\% & 16.7\% \\
Uncal+App    & 26.7\% & 3.3\%  & 6.7\%  & 12.2\% \\
Cal+App      & 16.7\% & 13.3\% & 13.3\% & 14.4\% \\
\bottomrule
\end{tabular}
\caption{Per-seed accuracy for MiniMax-M2.1 on GAIA.}
\label{tab:minimax-gaia-seeds}
\end{table}

\subsection{ALFWorld (202 tasks)}

\begin{table}[h]
\centering
\begin{tabular}{lcccc}
\toprule
Condition & Seed 42 & Seed 123 & Seed 456 & Mean \\
\midrule
Baseline        & 5.9\%  & 5.4\%  & 5.9\%  & 5.8\%  \\
Uncal+Roll      & 6.9\%  & 8.9\%  & 7.9\%  & 7.9\%  \\
Cal+Roll        & 6.9\%  & 8.9\%  & 5.0\%  & 6.9\%  \\
Uncal+App       & 7.9\%  & 7.4\%  & 10.4\% & 8.6\%  \\
Cal+App         & 7.9\%  & 6.4\%  & 8.9\%  & 7.8\%  \\
Late-only Roll  & 4.5\%  & 4.0\%  & 4.5\%  & 4.3\%  \\
Late-only App   & 6.9\%  & 5.0\%  & 5.4\%  & 5.8\%  \\
\bottomrule
\end{tabular}
\caption{Per-seed accuracy for Qwen-3-8B on ALFWorld (202 tasks).}
\label{tab:qwen-alf-seeds}
\end{table}

\begin{table}[h]
\centering
\begin{tabular}{lcccc}
\toprule
Condition & Seed 42 & Seed 123 & Seed 456 & Mean \\
\midrule
Baseline   & 13.9\% & 16.3\% & 13.9\% & 14.7\% \\
Uncal+Roll & 16.3\% & 16.3\% & 14.9\% & 15.8\% \\
Cal+Roll   & 14.9\% & 14.4\% & 15.8\% & 15.0\% \\
\bottomrule
\end{tabular}
\caption{Per-seed accuracy for GLM-4.7 on ALFWorld (202 tasks). GLM baseline failure rate (85.3\%) exceeds the threshold (74\%), and all conditions show positive or neutral effects, consistent with the framework's prediction.}
\label{tab:glm-alf-seeds}
\end{table}

\section{Qualitative Trajectory Examples}
\label{app:qual}

To illustrate the mechanisms discussed in the main paper, we present representative examples from Qwen-3-8B on HotPotQA (seed~42).

\paragraph{Example selection criteria.} Examples were selected to illustrate three primary mechanisms: (1)~early-step disruption, (2)~confidence erosion, and (3)~intervention cascades. We include both disruption and recovery cases. From seed~42, we observed 11 disruptions and 7 recoveries; we present 3 and 2, respectively, chosen for clarity of mechanism illustration.

\subsection{Disruption Examples (Baseline Success $\to$ Intervention Failure)}

\paragraph{Example 1: Immediate answer disrupted.}
\emph{Question:} ``What Cantonese slang term can mean both `ghost man' and to refer to Westerners?''
\emph{Ground truth:} Gweilo.
The baseline agent answered correctly at step~0. Under intervention, three LLM critic triggers caused the agent to second-guess its answer, ultimately returning Chinese characters instead of the requested English term. Steps: 0 (baseline) vs.\ 5 (intervention). Interventions: 3.

\paragraph{Example 2: Confidence erosion.}
\emph{Question:} ``Live Wire Radio will possibly be considered as a replacement for a variety show created in what year?''
\emph{Ground truth:} 1974.
The agent's immediate correct answer (1974) was disrupted by three interventions, causing it to search extensively and arrive at a wrong year (2004). Steps: 0 vs.\ 6. Interventions: 3.

\paragraph{Example 3: Strategy derailment.}
\emph{Question:} ``Did John Updike and Tom Clancy both publish more than 15 bestselling novels?''
\emph{Ground truth:} yes.
A single intervention caused the agent to shift from answering to outputting a code expression (\texttt{updike\_num > 15 and clancy\_num > 15}), resulting in a format error despite correct reasoning. Steps: 0 vs.\ 2. Interventions: 1.

\subsection{Recovery Examples (Baseline Failure $\to$ Intervention Success)}

\paragraph{Example 1: Factual correction.}
\emph{Question:} ``Who has released more solo albums, Nick Carter or Brady Seals?''
\emph{Ground truth:} Brady Seals.
The agent's immediate intuition (Nick Carter) was wrong. Interventions prompted additional research that led to the correct answer. Steps: 0 vs.\ 5. Interventions: 3.

\paragraph{Example 2: Knowledge retrieval.}
\emph{Question:} ``For One Night Only was hosted by the man most well-known for hosting what show from 1962 until 1999?''
\emph{Ground truth:} The Late Late Show.
The baseline answer was a plausible guess (Saturday Night Live) but wrong. Intervention-triggered research found the correct show (hosted by Gay Byrne). Steps: 0 vs.\ 6. Interventions: 3.

\subsection{Cascade Example}

\emph{Question:} ``Which `Roseanne' star is in Scream 2?''
\emph{Ground truth:} Laurie Metcalf.
The agent knew the answer immediately. The LLM critic intervened at step~0, triggering a search. The search action itself triggered another intervention (step~2). The cycle continued until the agent exhausted its step budget (18 steps) without calling \texttt{final\_answer}, producing no output despite having the correct answer initially. Steps: 0 vs.\ 18. Interventions: 3.

\subsection{Summary Statistics}

From Qwen-3-8B on HotPotQA (seed~42, 100 tasks):

\begin{table}[h]
\centering
\begin{tabular}{lcc}
\toprule
Category & Count & Notes \\
\midrule
Disruptions        & 11 & Baseline success $\to$ intervention failure \\
Recoveries         & 7  & Baseline failure $\to$ intervention success \\
Cascades ($\ge$3 interventions) & 78 & 78\% of tasks \\
Net effect         & $-4$ tasks & 11 disrupted $-$ 7 recovered \\
\bottomrule
\end{tabular}
\caption{Summary of disruption, recovery, and cascade statistics for Qwen-3-8B on HotPotQA (seed~42).}
\label{tab:qual-summary}
\end{table}

\section{Statistical Testing and Power Analysis}
\label{app:stats}

\paragraph{Bootstrap testing.}
All significance tests use paired, task-level bootstrap resampling (10{,}000 iterations). For each resample, we compute the difference in success rate between baseline and intervention on matched tasks and seeds. One-sided $p$-values test whether intervention improves performance. This procedure preserves task pairing, handles non-normal success distributions, and makes no parametric assumptions.

\paragraph{Multiple comparisons.}
For each benchmark--model pair, Holm--Bonferroni correction is applied across intervention conditions with family-wise $\alpha=0.05$. Uncorrected $p$-values are reported in tables; corrected significance is noted where applicable.

\paragraph{Power analysis.}
Table~\ref{tab:power} reports the minimum detectable effect size at 80\% power ($\alpha=0.05$).

\begin{table}[h]
\centering
\begin{tabular}{lcc}
\toprule
Benchmark & Tasks $\times$ Seeds & Detectable Effect \\
\midrule
HotPotQA & $100 \times 3$ & $\pm 5$ pp \\
GAIA & $30 \times 3$ & $\pm 12$ pp \\
ALFWorld & $202 \times 3$ & $\pm 4$ pp \\
\bottomrule
\end{tabular}
\caption{Statistical power by benchmark. Negative effects in high-success regimes are well above the detection threshold; positive effects in ALFWorld are near the resolution limit.}
\label{tab:power}
\end{table}

\section{Recovery and Disruption Accounting}
\label{app:dr}

We define recovery and disruption by comparing paired outcomes from matched baseline and intervention runs.

\paragraph{Definitions.}
For each task, we compare the baseline and intervention trajectories run from identical initial conditions:
\begin{itemize}
\item \textbf{Recovery}: the baseline trajectory fails, but the intervention trajectory succeeds.
\item \textbf{Disruption}: the baseline trajectory succeeds, but the intervention trajectory fails.
\end{itemize}

Rates are computed \emph{per task} (matched pairs), consistent with the definitions of $r = C/F$ and $d = B/S$ in Section~\ref{sec:framework}.

\paragraph{Aggregated counts.}
Table~\ref{tab:dr_counts} reports aggregated statistics across HotPotQA and GAIA.

\begin{table}[h]
\centering
\begin{tabular}{lcccccc}
\toprule
Model & $F$ & $C$ & $r$ & $S$ & $B$ & $d$ \\
\midrule
GLM-4.7 & 234 & 58 & 0.25 & 89 & 13 & 0.15 \\
Qwen-3-8B & 387 & 66 & 0.17 & 142 & 31 & 0.22 \\
MiniMax-M2.1 & 512 & 61 & 0.12 & 178 & 62 & 0.35 \\
\bottomrule
\end{tabular}
\caption{Recovery and disruption counts using notation from Section~\ref{sec:framework}: $F$ = baseline failures, $S$ = baseline successes, $C$ = recoveries, $B$ = disruptions, $r = C/F$, $d = B/S$. The ordering $(d/r)_{\text{MiniMax}} \gg (d/r)_{\text{Qwen}} > (d/r)_{\text{GLM}}$ is robust to bootstrap resampling.}
\label{tab:dr_counts}
\end{table}

\paragraph{Robustness.}
Counting raw intervention events (rather than collapsing to one per task) preserves the same ordering and yields similar thresholds, confirming that results are not an artifact of cascades.

\section{Early-Step Disruption Analysis}
\label{app:early}

To identify the mechanism behind negative effects in high-success regimes, we analyze all cases where baseline succeeds but intervention fails.

\paragraph{Definition.}
An episode is \emph{early-correct disrupted} if the baseline agent succeeds within 0--1 steps and the intervention run fails.

\paragraph{Results.}
Across all models, benchmarks, and mechanisms:

\begin{table}[h]
\centering
\begin{tabular}{lc}
\toprule
Failure Type & Fraction \\
\midrule
Early-step disruption (0--1 steps) & \textbf{100\%} \\
Mid-trajectory disruption & 0\% \\
\bottomrule
\end{tabular}
\caption{All observed regressions arise from interfering with trajectories that were already correct at steps 0--1.}
\label{tab:early-step}
\end{table}

\paragraph{Implication.}
This explains why (i)~calibration helps primarily by suppressing early triggers, (ii)~a minimum-step rule recovers 2--3 pp, and (iii)~intervention harm is concentrated in high-success regimes.

\section{Intervention Cascades and No-Answer Failures}
\label{app:cascades}

\paragraph{Cascades.}
An intervention cascade occurs when an intervention increases the likelihood of subsequent interventions within the same episode.

\begin{table}[h]
\centering
\begin{tabular}{lcc}
\toprule
Model & Cascade Rate & Avg.\@ Interventions \\
\midrule
GLM-4.7 & 91\% & 1.3 \\
Qwen-3-8B & 91\% & 2.5 \\
MiniMax-M2.1 & \textbf{96\%} & \textbf{3.1} \\
\bottomrule
\end{tabular}
\caption{Intervention cascade statistics by model on HotPotQA.}
\label{tab:cascades}
\end{table}

\paragraph{No-answer failures.}
On MiniMax / HotPotQA, the no-answer rate increases from 3.3\% at baseline to 46.1\% under intervention. Cascades exhaust the step budget, preventing final answer emission.

\paragraph{Interpretation.}
Cascades amplify the disruption--recovery imbalance. When $d>r$, repeated intervention compounds harm; when $r>d$, cascades compound benefit.

\section{Model Sensitivity Analysis}
\label{app:sensitivity}

MiniMax-M2.1 shows catastrophic sensitivity to intervention ($-$26 pp on HotPotQA) while GLM-4.7 shows near-neutral effects (0.0 pp to $-$3.7 pp). Two factors account for this difference.

\paragraph{Factor 1: Calibration controls intervention rate differently per model.}
Temperature scaling reduces the LLM critic's trigger rate, but the magnitude of this reduction varies across models (Table~\ref{tab:cal-rate}).

\begin{table}[h]
\centering
\begin{tabular}{lcccc}
\toprule
Model & Fitted $T$ & Uncal Int/Task & Cal Int/Task & Reduction \\
\midrule
GLM-4.7 & 8.81 & 2.50 & 0.73 & $-$71\% \\
MiniMax-M2.1 & 1.05 & 2.78 & 2.71 & $-$3\% \\
Qwen-3-8B & 2.27 & 2.36 & 1.74 & $-$26\% \\
\bottomrule
\end{tabular}
\caption{Calibration effect on intervention rate. GLM's high fitted temperature (8.81) flattens the LLM critic's probability distribution, reducing interventions by 71\%. MiniMax's near-unity temperature (1.05) provides almost no reduction.}
\label{tab:cal-rate}
\end{table}

Calibration helps GLM not by improving prediction accuracy but by making the LLM critic less trigger-happy. MiniMax receives nearly every intervention the LLM critic wants to fire.

\paragraph{Factor 2: Recovery rate varies across models.}
When interventions do occur, models differ in how often they recover (Table~\ref{tab:recovery-rate}).

\begin{table}[h]
\centering
\begin{tabular}{lcc}
\toprule
Model & Recovery Rate & Interpretation \\
\midrule
GLM-4.7 & 25\% & 1 in 4 interventions leads to recovery \\
Qwen-3-8B & 17\% & 1 in 6 interventions leads to recovery \\
MiniMax-M2.1 & 12\% & 1 in 8 interventions leads to recovery \\
\bottomrule
\end{tabular}
\caption{Per-intervention recovery rates by model. GLM recovers twice as often as MiniMax.}
\label{tab:recovery-rate}
\end{table}

Combined, these factors produce MiniMax's 7.3:1 disruption-to-recovery ratio (vs.\ 1.5:1 for GLM). For every task MiniMax recovers, 7.3 are disrupted.

\paragraph{No-answer effect.}
Intervention cascades can prevent agents from producing any answer at all. On MiniMax/HotPotQA, the no-answer rate increases from 3.3\% at baseline to 46.1\% under intervention (a 14$\times$ increase). The agent becomes trapped in rollback loops and exhausts its step budget before calling \texttt{final\_answer}. This no-answer failure mode accounts for a substantial fraction of MiniMax's performance collapse.

\paragraph{Implication.}
The same LLM critic and threshold can cause moderate or catastrophic harm depending on how calibration interacts with model-specific characteristics. Successful intervention requires controlling intervention rate, not just prediction accuracy.

\section{Oracle Bounds and the Disruption Tax}
\label{app:oracle}

To decouple prediction quality from intervention mechanics, we compute oracle upper bounds.

\paragraph{Oracle intervention.}
Intervene only on trajectories that fail under baseline.

\paragraph{Oracle selection.}
Choose the better of two completed trajectories (Best-of-2).

\begin{table}[h]
\centering
\begin{tabular}{lccc}
\toprule
Model & Baseline [\%] & Oracle Interv.\ [\%] & Oracle Bo2 [\%] \\
\midrule
Qwen-3-8B & 57.0 & 64.7 (+7.7) & 68.0 (+11.0) \\
GLM-4.7 & 70.3 & 75.0 (+4.7) & 77.0 (+6.7) \\
MiniMax-M2.1 & 64.0 & 68.0 (+4.0) & 75.0 (+11.0) \\
\bottomrule
\end{tabular}
\caption{Oracle ceilings on HotPotQA. Post-hoc selection has substantially higher headroom than mid-execution intervention.}
\label{tab:oracle}
\end{table}

\paragraph{Disruption tax.}
The gap between oracle selection and oracle intervention (6--11 pp) quantifies an intrinsic disruption tax imposed by mid-execution control. Improving prediction accuracy alone cannot eliminate this gap.

\section{Exploratory Analysis: LLM critic-Based Selection}
\label{app:guard}

We evaluate whether LLM critic scores can act as a trajectory-level selection signal. These results are preliminary due to limited statistical power and are intended to motivate future work rather than to support confirmatory claims. For contested tasks (one seed succeeds, one fails), we select the trajectory with lower LLM critic score (predicting success).

\begin{table}[h]
\centering
\begin{tabular}{llccccc}
\toprule
Model & Benchmark & Contested & Oracle $\Delta$ & LLM critic $\Delta$ & LLM critic Acc & Judge Acc \\
\midrule
GLM   & HotPotQA & 11 & +5.6 pp & +1.5 pp & 63.6\% & 61.1\% \\
Qwen  & HotPotQA & 8  & +5.3 pp & $-$2.6 pp & 25.0\% & 66.7\% \\
\bottomrule
\end{tabular}
\caption{LLM critic-based selection on contested tasks. With $n=11$ contested tasks, statistical power is limited (23\% to detect a 15 pp accuracy difference at $\alpha=0.05$).}
\label{tab:guard-selection}
\end{table}

\paragraph{Interpretation.}
Oracle headroom is robust (+11 pp average), motivating selection-based approaches. Answer-only LLM judgment fails (51.7\% $\approx$ random). LLM critic-based selection requires larger-scale validation; a properly powered study would need $n \ge 50$ contested tasks for 80\% power.

\section{Boundary Condition Analysis: Non-Agentic Settings}
\label{app:boundary}

The disruption--recovery framework assumes multi-step agentic execution where agents take sequential actions, intervention occurs mid-trajectory, and recovery is possible ($r > 0$). We test what happens when these assumptions are violated using single-shot code generation.

\paragraph{Prediction.}
When $r \to 0$, the threshold $p^\star = d/(r+d) \to 1$. Intervention should yield neutral to negative effects regardless of baseline failure rate.

\paragraph{Setup.}
We use SWE-bench Lite \cite{jimenez2024swebench} with single-shot patch generation (no iterative debugging, no tool use). Models: Qwen-3-8B, MiniMax-M2.1. Tasks: 30 instances $\times$ 3 seeds.

\begin{table}[h]
\centering
\begin{tabular}{llccccc}
\toprule
Model & Condition & Seed 42 & Seed 123 & Seed 456 & Mean & $\Delta$ \\
\midrule
Qwen-3-8B & Baseline & 10.0\% & 10.0\% & 10.0\% & 10.0\% & -- \\
Qwen-3-8B & Rollback & 6.7\%  & 10.0\% & 6.7\%  & 7.8\%  & $-$2.2 pp \\
Qwen-3-8B & Append   & 10.0\% & 10.0\% & 10.0\% & 10.0\% & 0.0 pp \\
\bottomrule
\end{tabular}
\caption{SWE-bench Lite results (non-agentic single-shot generation). Consistent with the framework's prediction for $r \to 0$: intervention yields neutral to marginally negative effects.}
\label{tab:swebench}
\end{table}

\paragraph{Analysis.}
Intervention yields neutral (\texttt{APPEND}: 0.0 pp) to marginally negative (\texttt{ROLLBACK}: $-$2.2 pp) effects for Qwen. Despite a high failure rate ($p = 90\%$), the framework correctly predicts that without recovery dynamics, intervention cannot help. MiniMax achieves 0\% across all conditions, reflecting insufficient base capability for this task format.

\paragraph{Retry is not recovery.}
In agentic settings, rollback preserves partial progress and the agent can adapt its strategy. In single-shot settings, each attempt is statistically independent; the intervention provides no information the model can use. This distinction confirms that the framework's assumptions are appropriately scoped to multi-step agentic execution.

\paragraph{Boundary validation.}
Observing neutral-to-negative effects in a setting where the framework predicts intervention should not help provides additional evidence for the framework's validity. The disruption--recovery criterion produces testable predictions that hold both within the intended operating regime and at the boundary where its assumptions are violated.

\paragraph{Statistical note.}
With 30 tasks $\times$ 3 seeds $=$ 90 observations per condition, we have 80\% power to detect effects $\ge$12 pp (two-sided $\alpha = 0.05$). The observed Qwen effects ($-$2.2 pp, 0.0 pp) are within the noise floor. We interpret these as consistent with a neutral effect rather than as precise effect estimates.

\section{Prompts}
\label{app:prompts}

\subsection{LLM Critic Feedback Prompt}
The following message is appended to the agent's context when the LLM critic triggers under the \texttt{APPEND} mechanism:
\begin{lstlisting}
The LLM critic model predicts this action may
lead to task failure. Please reconsider your
approach.
\end{lstlisting}

\subsection{Contextual LLM Feedback Prompt}
For the LLM-generated feedback ablation, the following template is used:
\begin{lstlisting}
STOP. The previous approach may lead to an
incorrect answer (predicted failure probability:
{failure_probability:.0%}). Please reconsider
and try a different approach. Think more
carefully about the question.
\end{lstlisting}

\end{document}